\documentclass[letterpaper, 10 pt, conference]{ieeeconf}  

\IEEEoverridecommandlockouts                              

\usepackage{amsmath,amsfonts}
\usepackage{amsthm}
\usepackage{array}
\usepackage{graphicx}
\usepackage{textcomp}
\usepackage{stfloats}
\usepackage{url}
\usepackage{verbatim}
\usepackage{amssymb}
\usepackage{pifont}
\usepackage{siunitx}
\usepackage{algorithm,algcompatible}
\usepackage{algpseudocode}
\usepackage{hyperref}
\usepackage[capitalize]{cleveref}
\usepackage{comment}
\usepackage{multirow}
\usepackage{booktabs}
\usepackage{subcaption}
\algnewcommand\INPUT{\item[\textbf{Input:}]}
\algnewcommand\OUTPUT{\item[\textbf{Output:}]}
\DeclareMathOperator*{\argmax}{\arg\!\max}

\usepackage[mode=buildnew]{standalone}

\title{\LARGE \bf A Safe Reinforcement Learning driven\\ Weights-varying Model Predictive Control\\for Autonomous Vehicle Motion Control}
\author{Baha Zarrouki, Marios Spanakakis and Johannes Betz
\thanks{B. Zarrouki,  M. Spanakakis and J. Betz are with the Professorship of Autonomous Vehicle Systems, TUM School of Engineering and Design, Technical University Munich; Munich Institute of Robotics and Machine Intelligence (MIRMI), \{{baha.zarrouki}, {m.spanakakis}, {johannes.betz}\}@tum.de}}
\begin{document}
\maketitle
\thispagestyle{empty} 
\pagestyle{empty}    
\begin{abstract}
Determining the optimal cost function parameters of Model Predictive Control (MPC) to optimize multiple control objectives is a challenging and time-consuming task. Multi-objective Bayesian Optimization (BO) techniques solve this problem by determining a Pareto optimal parameter set for an MPC with static weights. However, a single parameter set may not deliver the most optimal closed-loop control performance when the context of the MPC operating conditions changes during its operation, urging the need to adapt the cost function weights at runtime. 
Deep Reinforcement Learning (RL) algorithms can automatically learn context-dependent optimal parameter sets and dynamically adapt for a Weights-varying MPC (WMPC). However, learning cost function weights from scratch in a continuous action space may lead to unsafe operating states.
To solve this, we propose a novel approach limiting the RL actions within a safe learning space representing a catalog of pre-optimized BO Pareto-optimal weight sets. We conceive a RL agent not to learn in a continuous space but to proactively anticipate upcoming control tasks and to choose the most optimal discrete actions, each corresponding to a single set of Pareto optimal weights, context-dependent. Hence, even an untrained RL agent guarantees a safe and optimal performance.
Experimental results demonstrate that an untrained RL-WMPC shows Pareto-optimal closed-loop behavior and training the RL-WMPC helps exhibit a performance beyond the Pareto-front.
\end{abstract}



\section{Introduction}
The performance of an MPC heavily relies on the formulation of its cost function, especially the appropriate selection of cost function weights, a pivotal aspect of MPC design.
This process poses three key challenges: it demands significant manual effort and domain expertise, making it time-consuming and susceptible to suboptimality; the cost function profoundly influences operational safety, as unsuitable weight sets render the control problem infeasible; and a given weight distribution is valid only for a specific operating point, with changes in operating conditions potentially causing substantial performance degradation. This safety concern is particularly relevant for systems like autonomous vehicles, operating across diverse conditions.
Approaches in the literature address these issues through controller design automation. In \cite{ramasamy2019optimal}, multi-objective Genetic Algorithms (GA) are used to determine MPC weights, enhancing closed-loop performance. Additionally, \cite{rodrigues2019tuning} presents a GA approach that handels constraints. 
\begin{figure}[H]
\centering
    \begin{subfigure}[t]{1\columnwidth}
    \includegraphics[width=1\columnwidth]{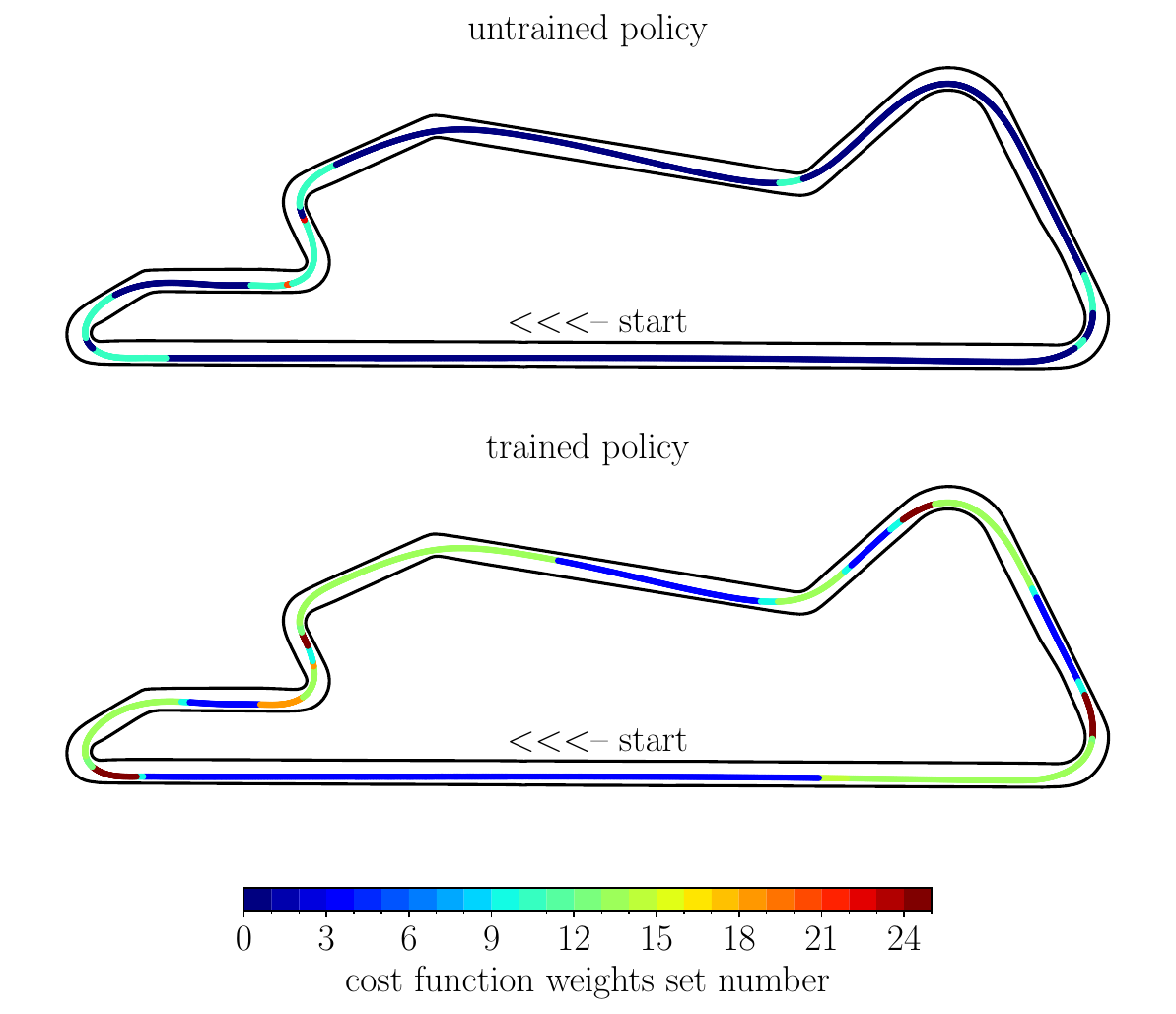}
    \caption{The RL agent dynamically adjusts MPC cost function weights online to align with an optimal raceline trajectory on the Monteblanco racetrack. The RL agent selects a set from a pre-computed catalog of Pareto optimal weights for each switching time. The sets (0-10) are optimized for straight segments, (15-25) for curves, and (11-14) strike a tradeoff between the two.}
    \label{fig:heatmap}
    \end{subfigure}
    \newline
    \begin{subfigure}[t]{1\columnwidth}
    \includegraphics[width=1\columnwidth]{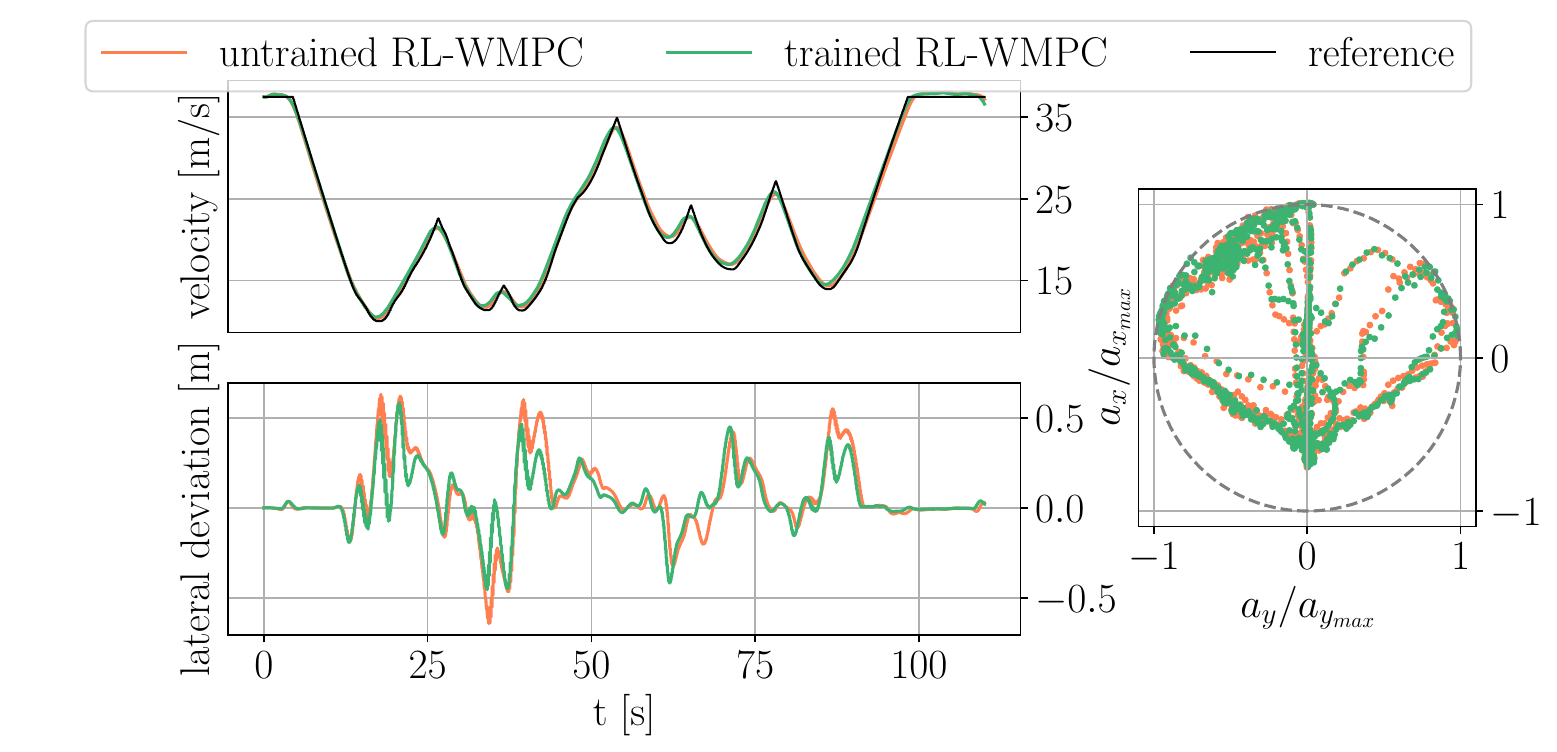}
    \caption{Velocity-, lateral deviation and the gg-diagram plots.}
    \label{fig:benchmark_trained_untrained_gg}
    \end{subfigure}
    
\caption{Safe RL driven Weights-varying MPC: comparison between an untrained- and a trained policy. Even the untrained RL policy shows good closed-loop behavior, attributed to the global safety inherent in the pre-computed weights catalog obtained through Multi-objective Bayesian Optimization (MOBO).}
\label{fig:trained_vs_untrained}
\end{figure}
Furthermore, collaborative Particle Swarm Optimization (PSO) are employed in several instances to navigate the search space and identify tuning parameters that enhance trajectory tracking performance, particularly in the context of MPCs controlling quadrotors \cite{kapnopoulos2022cooperative}.
Another class of algorithms employed for automatic MPC tuning falls under Bayesian Optimization (BO). \cite{piga2019performance} and \cite{gharib2021multi} propose performance-driven MPC tuning strategies using BO, while \cite{sorourifar2021data}, \cite{paulson2021probabilistically}, and \cite{rupenyan2021performance} utilize constrained BO in their respective works. Additionally, \cite{makrygiorgos2022performance} puts forth a multi-objective BO approach to mitigate plant-model mismatch.
Moreover, Reinforcement Learning (RL) proves valuable in learning controller design. In \cite{zarrouki2023adaptive}, an approach is presented to automatically tune Stochastic MPC meta-parameters online using Deep RL, enhancing uncertainty handling, constraints robustification, feasibility, and closed-loop performance. \cite{bohn2021reinforcement} employs RL algorithms to determine the prediction horizon of MPC, reducing computational complexity, while \cite{bohn2023optimization} focuses on learning other MPC meta-parameters. Additionally, \cite{mehndiratta2018automated} introduces RL for the automatic tuning and determination of a static set of weights for an MPC controlling a quadrotor.
Meanwhile, \cite{zarrouki2021weights}\cite{zarrouki2020reinforcement} present a Weights-varying Nonlinear MPC driven by Deep RL, enabling the automatic learning and online adaptation of cost function weighting matrices to optimize multi-control objectives. However, the challenge arises when learning cost function weights in a continuous action space, as safety is not inherently guaranteed. Our approach tackles this concern by constraining the action space to discrete options, allowing the agent to select sets from a catalog of pre-optimized weights determined by Bayesian Optimization (BO).
In the study by \cite{wabersich2020performance}, safety considerations are addressed when integrating Reinforcement Learning (RL) with Model Predictive Control (MPC) within a resulting Bayesian-MPC framework. In \cite{berkenkamp2016safe}\cite{berkenkamp2017safe}, the emphasis is on optimizing control law parameters with a safety guarantee. Their approach models the performance measure as a Gaussian process and explores new controller parameters, specifically those with a high probability of performance above a safe threshold. However, these works do not specifically tackle finding optimal cost function weights.

In summary, our work contributes in three key ways:
\begin{itemize}
    \item 
    We conceive a Weights-varying MPC (WMPC) driven by a Safe RL agent. The RL agent uses a look-ahead design to adjust MPC cost function weights autonomously under varying operating conditions, selecting the most suitable set from a Pareto front optimized through Multiobjective Bayesian Optimization (MOBO).
    \item We demonstrate the adaptability of our approach in dynamically adjusting the weights of a Nonlinear MPC for controlling a full-scale autonomous vehicle to follow an optimal raceline. Additionally, we provide safety evidence, showcasing Pareto optimality even with an untrained agent. 
    \item We perform a context-dependent analysis of the RL agent's decision-making process and assess our approach's generalization and robustness capabilities through testing in unseen environments. We also explore the impact of continuous learning in an inexperienced environment.
\end{itemize}
\section{Weights-varying MPC}\label{sec:meth}

Weights-varying MPC (WMPC) \cite{zarrouki2021weights} is an MPC that adapts its cost function weights during online operation. Different weights optimized for several operation conditions enhance the overall closed-loop performance and show advantages compared to an MPC operating with a static cost function weights set. A Weights scheduler module provides the MPC with a new set of weights computed by a weights optimization policy. The Weights scheduler can be designed to update the MPC event-based or following a schedule. 
\begin{figure}[H]
  \centering
  \includegraphics[width=1\columnwidth]{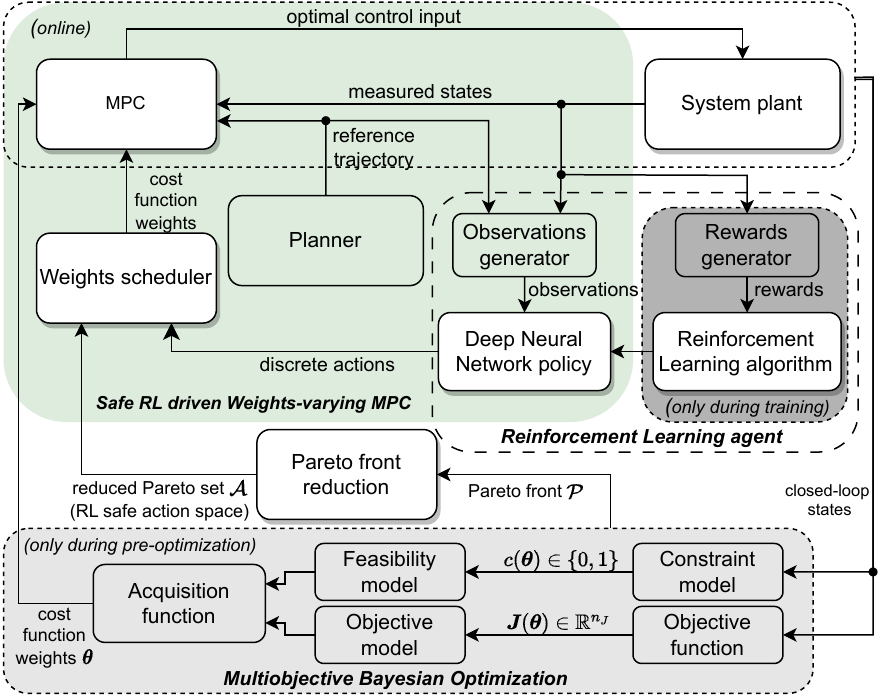}
  \caption{Architecture of the Safe RL driven Weights-varying MPC}
  \label{fig: architecture of DRL-WMPC}
\end{figure}
In this work, we advocate a Deep Neural Network (DNN) policy to determine optimal cost function weights following a fixed schedule based on observations of relevant closed-loop states and future MPC reference trajectories. We conceive the DNN policy to choose every switching time $T_{sw}$ a cost function weights set from a catalog of pre-optimized safe weights. This catalog represents a Pareto set $\mathcal{A}$ that we determine with an offline Multi-Objective Bayesian Optimization (Sec.\ref{sec:pareto_sets}). The set $\mathcal{A}$ contains cost function weights locally optimized for single operating conditions. Yet, each set is valid for a global safe operation, i.e., for an MPC with static weights.\\
During training, a DNN policy offline is learned with a Deep Reinforcement Learning (RL) algorithm (Sec.\ref{sec:rl_snmpc}). Each RL action is safe as the agent does not learn cost function weights, but automatically learns which optimal cost function weights from the catalog $\mathcal{A}$ are optimal depending on the context and when to switch them. Thus, we conceive a Safe Deep RL driven WMPC. Figure \ref{fig: architecture of DRL-WMPC} denotes its architecture.
\section{Static Nonlinear MPC}
We consider the following static Nonlinear MPC (NMPC) where, $\boldsymbol{x} \in \mathbb{R}^{n_x}$ denotes the state vector, $\boldsymbol{u} \in \mathbb{R}^{n_u}$ the control vector, $t$ the discrete time, $T_p$ the prediction horizon, $f$ the system dynamics, $\boldsymbol{h}$ and $\boldsymbol{h}^e$ the path and terminal inequality constraints on states and control inputs, which can be linear or nonlinear, and $\boldsymbol{x}_0$ the initial state.
\begin{equation}
\begin{aligned}
& \textbf{Problem 1} && \textbf{Nominal NMPC}\\ 
&  \underset{\boldsymbol{x}(.), \boldsymbol{u}(.)}{\min} & & 
\begin{aligned}
    \int^{T_p}_{\tau=0} & l(\boldsymbol{x}(\tau),\boldsymbol{u}(\tau)) \space  d\tau \\ 
 & + m(\boldsymbol{x}(T_p))
\end{aligned}
  \\ 
& \text{subject to} & & \boldsymbol{x}_{0} \leq \boldsymbol{x}(0) \leq \boldsymbol{x}_{0} \text {, }  \\
& & & \dot{\boldsymbol{x}}(t) = f(\boldsymbol{x}(t),\boldsymbol{u}(t)) \text {, } & t \in[0, T_p), \\
& & &  \boldsymbol{h}(\boldsymbol{x}(t), \boldsymbol{u}(t)) \leq \bar{\boldsymbol{h}},& t \in[0, T_p), \\
& & &   \boldsymbol{h}^{\mathrm{e}}(\boldsymbol{x}(T_p)) \leq \bar{\boldsymbol{h}}^{\mathrm{e}}, \\
\end{aligned}
\label{eq:nominal NMPC problem}
\end{equation}
For the cost function, we employ a nonlinear least squares approach for the stage cost  $l(\boldsymbol{x}, \boldsymbol{u})=\frac{1}{2}\|\boldsymbol{y}(\boldsymbol{x},\boldsymbol{u})-\boldsymbol{y}_{\mathrm{ref}}\|_W^2$, where $l: \mathbb{R}^{n_{\mathrm{x}}} \times \mathbb{R}^{n_{\mathrm{u}}}  \rightarrow \mathbb{R}$, and for the terminal cost $m(\boldsymbol{x})=\frac{1}{2}\|\boldsymbol{y}^e(\boldsymbol{x})-\boldsymbol{y}^e_{\mathrm{ref}}\|_{W^e}^2$, where $m: \mathbb{R}^{n_{\mathrm{x}}}  \rightarrow \mathbb{R}$. Here, $W$ and $W_e$ represent weighting matrices for the stage and terminal costs, respectively. Specifically, $W$ is computed as $W = \text{diag}(Q,R)$, where $Q$ and $R$ are matrices used for weighting states and inputs, and $W_e = Q_e$. \\
In this work, we consider the combined longitudinal and lateral motion control of a full-scale autonomous passenger vehicle. For the full MPC problem formulation, we refer to work \cite{zarrouki2023stochastic}. The weighting matrices are defined as follows: 
\begin{equation}
  \begin{aligned}
Q &= \operatorname{diag}(q_{x,y}, q_{x,y}, q_\psi, q_v) \\
R &= \operatorname{diag}(r_j, r_\omega)\\
\end{aligned}  
\end{equation}
Here, $q_{x,y}$ denote the weight for the x- and y-coordinates deviation of the ego vehicle from the reference, $q_\psi$ and $q_v$ the yaw angle and velocity deviation respectively, $r_j$ the weight for the longitudinal jerk, and $r_\omega$ the steering rate weight. Moreover, we employ two slack variables, $L1$ for a linear- and $L2$ for a quadratic constraint violation penalization term, relaxing the constraints and helping the solver find a solution.\\
This work aims to automatically learn and adapt the weighting matrices, Q and R, and the slack terms $L1$ and $L2$. 
\section{Multi-objective Bayesian Optimization:
Determining Pareto Optimal Weight Sets}%
\label{sec:pareto_sets}%
We require a safe action space to allow for Safe RL as described above. This is obtained by identifying optimal, viable weight sets within the defined parameter space. The tuning task is formulated as a multiobjective optimization problem and solved using Multiobjective Bayesian Optimization (MOBO) to obtain the set of Pareto optimal weight sets.
\subsection{Formulating the optimization problem}%
For the given control design task, the decision variables are the cost function weights and soft constraint penalty terms:
\begin{equation}
\begin{aligned}
\boldsymbol{\theta} &= [q_{x,y},\;q_{\psi},\;q_{v},\;r_{j},\;r_{\omega},\;L_1,\;L_2]\\
\end{aligned}
\end{equation}
In the optimization process, the decision variables are fine-tuned to minimize two distinct performance metrics: the vehicle's lateral deviation from the reference trajectory and the deviation from the reference velocity. Specifically, the lateral deviation metric focuses on the maximum observed deviation throughout the entire trajectory:
\begin{equation}
\begin{aligned}
J_0(\boldsymbol{\theta}) &= \max \left( e_\text{lat, 0}, e_\text{lat, 1}, \ldots, e_\text{lat, N} \right)
\end{aligned}
\end{equation}
here, $e_\text{lat,k}$ denotes the lateral deviation at step $k$. The velocity deviation metric is defined as:
\begin{equation}
\begin{aligned}
J_1(\boldsymbol{\theta}) &= \sqrt{\frac{1}{N}\sum_{k=0}^{N} \left( v_{k}(\boldsymbol{\theta}) - v_{ref, k} \right) ^{2}}\,.
\end{aligned}
\end{equation}
This metric is formulated as a root-mean-square (RMS) error. 
To ensure solution feasibility, i.e., the compliance to physical constraints and the absence of intolerable behavior, each solution must satisfy two global constraint conditions: 
\begin{enumerate}
    \item A feasible parameter set must not violate the MPC's constraints, e.g., maximum allowed combined lateral and longitudinal accelerations.
    \item The lateral deviation must not exceed a defined maximum value $e_\text{lat, max}$.
\end{enumerate}
Consequently, we formulate the optimization problem as:
\begin{equation}
    \begin{aligned}
    & \textbf{Problem 2} && \textbf{Multiobjective Bayesian Optimization} \\ 
        &\min_{\boldsymbol{\theta}\in\mathbb{R}^{n_\theta}} &&\boldsymbol{J}(\boldsymbol{\theta})\;=\;[J_{1}(\boldsymbol{\theta}), \ldots, J_{n_J}(\boldsymbol{\theta})] \\
    &\text{subject to} \qquad &&\boldsymbol{\theta}_{lb} \le \boldsymbol{\theta} \le \boldsymbol{\theta}_{ub} \\
    &&& \boldsymbol{g}(\boldsymbol{\theta}) \leq \bar{\boldsymbol{g}} 
    \end{aligned}
    \label{eq: MOBO problem}
\end{equation}
where $\boldsymbol{J}$ denotes the vector containing the optimization objectives and $\boldsymbol{g}$ the optimization constraint functions. $\boldsymbol{\theta}_{lb}$ and $\boldsymbol{\theta}_{ub}$ denote the individual boundaries for the decision variables.
\subsection{Surrogate models}%
The surrogate models approximate the real, unknown objective function and serve to select the most promising point to sample next, i.e., a parameter set that is likely to be beneficial for finding global optima. The surrogate model is constructed from a set of three Gaussian Processes (GPs) for the given system. Two of the GPs model the expected outcome for each optimization objective. The third GP is a binary classifier that models the probability of feasibility for each point in the decision space. This feasibility model utilizes the information gained from system failures and crashes due to unfavorable weight sets. 
Each model uses a constant mean function and the Radial Basis Function (RBF) kernel. 
The third GP model transforms the classification task into a regression task by leveraging a modified Dirichlet likelihood. 
Given a configuration $\theta$, the model outputs a probability distribution describing the expected solution feasibility. 
\subsection{Acquisition function}%
The acquisition function encapsulates the strategy for selecting promising sample points, which is decisive for the sample efficiency of BO. Various acquisition functions have been proposed for multiobjective optimization, of which the most established is the Expected Hypervolume Improvement (EHVI) \cite{Emmerich2006}. This function bases the search strategy on the volume in objective space enclosed by the Pareto front and a reference point $R$. Essentially, the EHVI acquisition function serves as a decision-making criterion, favoring candidate solutions that offer the most significant expected improvement to the multidimensional objective space, thereby steering the optimization process towards regions with high potential for Pareto front expansion. The EHVI acquisition function itself does not provide means to include the feasibility estimate of the third surrogate model in the decision process. To take solution feasibility into account, the feasibility estimate obtained from the feasibility model is used as a weight for the acquisition function.\\
The acquisition value $\alpha(\vec{\theta})$ for any given $\vec{\theta}$, i.e., for any parameter space possible configuration, as well as the feasibility weight $\alpha_\text{feas}(\vec{\theta})$ are defined as:
\begin{equation}
    \begin{aligned}
        \alpha(\vec{\theta}) &= \alpha_\text{EHVI}(\vec{\theta}) \; \cdot \; \alpha_\text{feas}(\vec{\theta}) \\
        \alpha_\text{feas}(\vec{\theta}) &= \min \left( \hat{\mu}_\text{feas}\left(\vec{\theta}\right)^{k} \; + \; \epsilon \cdot \hat{\sigma}_\text{feas}(\vec{\theta}) , \; 1\right)
    \end{aligned}
\end{equation}
The latter is calculated based on $\hat{\mu}_\text{feas}(\vec{\theta})$ and $\hat{\sigma}_\text{feas}(\vec{\theta})$, which denote the mean and standard deviation of the feasibility estimate. 
The parameter $k$ provides a mean to control the degree of trust in the probability estimate interpretation. For $k > 1$, estimated probability values below 1 are skewed towards 0, encouraging sampling for highly likely feasible configurations. Conversely, for $k < 1$, smaller feasibility estimates are biased towards 1, increasing the likelihood of testing such configurations.
Scaling $\hat{\sigma}_\text{feas}(\vec{\theta})$ by $\epsilon$ offers control over the exploration degree and encourages sampling in regions lacking reliable feasibility estimates. All experiments have fixed the feasibility weight parameters at $k = 1$ and $\epsilon = 0.8$. It is important to note that the feasibility weight is capped at 1. While the weighting can reduce the likelihood of sampling unfeasible regions, it doesn't have additional influence on the selection process in regions expected to be feasible.
\subsection{Training track segmentation}%
To specialize parameterizations for distinct track features, our approach involves segmenting training tracks based on curvature profiles. MOBO evaluates control performance separately for each segment group, allowing the distinction between parameter sets excelling in curved sections and those performing well on straight-track segments. This results in two parallel optimization processes seeking the best performance on their respective trajectory types. Note that a specialized parameterization is considered, only if it respects the global feasibility condition.
\subsection{Optimization algorithm}%
The optimization framework concurrently optimizes for two scenarios using a parallel approach. Each segment group's parameterization is independently tested, producing two distinct datasets for performance on straight and curved sections. For the next iteration, a sample point is selected alternately from one of the datasets, and surrogate models are fitted, considering joint constraint fulfillment. The strategy in Algorithm \ref{alg:parallel_constrained_bayesian_optimization} enhances efficiency by sharing information between the two optimization processes, ensuring that a parameter set inducing unfeasible behavior in one scenario is deemed non-viable, regardless of performance in the other. This parallel search generates two closely connected Pareto fronts, facilitating the sharing of entries. 
\begin{algorithm}
\caption{Parallel constrained Bayesian optimization for multiple evaluation environments}
\label{alg:parallel_constrained_bayesian_optimization}
\begin{algorithmic}[1]
    \OUTPUT set of non-dominated solutions $\mathcal{P}$
    \State{obtain initial data set $\mathcal{D} \gets \{\vec{\theta}_{i}, \; \vec{J}(\vec{\theta}_{i}), \; c(\vec{\theta}_{i})\}$ for $i = 0, \ldots, n_\text{init}$}
    \For{$k \in (1, \ldots, N)$}
        \State{fit objective model to $\vec{J}(\vec{\theta}_{i})$}
        \State{fit constraint model to $c(\vec{\theta}_{i})$}
        \State{obtain next sample point $\vec{\hat{\theta}}_{k} \gets \argmax{\alpha(\vec{\theta})}$}
        \State{evaluate objective function $\vec{f}(\vec{\hat{\theta}}_{k})$}
        \State{augment data set $\mathcal{D}_{k} \gets \{\vec{\hat{\theta}}_{k}, \vec{J}(\vec{\theta}_{k}), \; c(\vec{\theta}_{k}) \}$}
    \EndFor
\end{algorithmic}
\end{algorithm}

\subsection{Pareto front reduction}%
The optimization process yields dense Pareto fronts. Solution clusters may emerge from convergence, impacting RL by favoring certain behaviors. To address this, reducing the number of actions is beneficial, preventing biases while maintaining strategy diversity. Balancing action pruning is crucial to avoid compromising the exploration of available parameterizations.
We employ a reduction technique based on relative distances in the objective space to condense the action space while maintaining diversity among behaviors (Algorithm \ref{alg:point_cloud_reduction}). We define the reduced Pareto front $\mathcal{A}$.
\begin{algorithm}
\caption{Pareto front reduction with k-means clustering}
\label{alg:point_cloud_reduction}
\begin{algorithmic}[1]
    \INPUT dense $n_J$ dimensional Pareto front composed of $n_{\mathcal{P}}$ points: $\mathcal{P} = \{ p_0, \ldots, p_{n_{\mathcal{P}-1}} \}$ with $p_i \in \mathbb{R}^{n_{\boldsymbol{\theta}}} \; \forall i \in [0, \ldots, n_{\mathcal{P}-1}]$
    \OUTPUT{reduced set of $n_\mathcal{A}$ equally-distributed Pareto points $\mathcal{A}= \{ \boldsymbol{\theta}^{(0)}, \ldots, \boldsymbol{\theta}^{(n_\mathcal{A}-1)} \}$  with $\boldsymbol{\theta}^{(i)} \in \mathbb{R}^{n_{\boldsymbol{\theta}}} \; \forall i \in [0, \ldots, n_\mathcal{A}-1)]$}
    \State min-max normalize $\mathcal{P}$
    \State $\mathcal{A} \gets \emptyset$ 
    \For{$i \in \left\{1,...,n_J\right\}$}
        \State{$p_{\text{min}, i} \gets \arg\min_{p \in \mathcal{P}}\{ p[i]\}$}
        \State{$\mathcal{A} \gets \mathcal{A} \cup \{ p_{\text{min}, i} \} $}
    \EndFor
    \State{$\{C_1, C_2, \dots, C_{n_\mathcal{A}-n_{\boldsymbol{\theta}}}\} \gets \text{k-Means}(\mathcal{P}, n_\mathcal{A}-n_J)$}
    \For{$i \in \left\{1,...,n_\mathcal{A} - n_J\right\}$}
        \State{$\mathcal{A} \gets \mathcal{A} \cup \{C_i\} $}
    \EndFor
\end{algorithmic}
\end{algorithm}
First, the minima along each dimension, representing the best global performance for each objective, are added to the set $\mathcal{A}$. Then, the remaining parameterizations are grouped using k-means clustering to extract a representative subset. This iterative process develops clusters with similar characteristics, and the center of each cluster, representing a behavior, is added to $\mathcal{A}$.
\section{Deep Reinforcement Learning: Safe Learning Weights-varying MPC}
\label{sec:rl_snmpc}
\begin{figure}[h]
  \centering
  \includegraphics[width=0.47\textwidth]{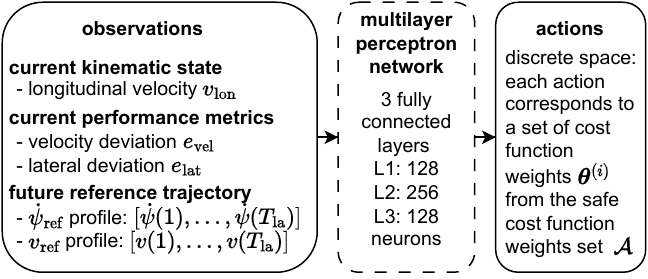}
  \caption{Architecture of the Deep Neural Network driven WMPC}
  \label{fig:NN_architecture}
\end{figure}

In this section, we introduce our Weights-varying MPC (WMPC), designed to dynamically adjust cost function weights for optimal closed-loop performance in response to changing environments and driving tasks. Utilizing a Deep Neural Network (DNN) policy trained through state-of-the-art Deep RL (DRL) algorithms, we illustrate the architecture of our DRL-driven WMPC in Fig.\ref{fig: architecture of DRL-WMPC}. The DNN policy computes new actions (Sec.\ref{subsec:action space}), and the parameter scheduler adjusts WMPC parameters at predefined switching time instances $T_{sw} = n\cdot T_\text{s,sim}, n \in \mathbb{N}$, where $T_\text{s,sim}$ is the environment discretization time. New actions are computed via a feed-forward step through the DNN using observation vector stacks generated from current measured states and the future reference trajectory (Sec.\ref{subsec:observation space}).\\
During the training/learning phase, the Proximal Policy Optimization (PPO) method \cite{schulman2017proximal} updates the DNN policy based on rewards collected (Sec.\ref{subsec: designing reward function}) after a specified number of steps in the environment $n_\text{steps}$, i.e., new updated WMPC parameters steps. Following training, the rewards generator and the RL algorithm are no longer necessary, and the architecture remains unchanged for deployment. Figure \ref{fig:NN_architecture} outlines the architecture of the DNN policy tailored for this problem.

\subsection{Defining the action space}
\label{subsec:action space}
We design a discrete action space. Each action $a$ corresponds to one set of cost function weights $\boldsymbol{\theta}^{(i)}$ from the pre-optimized Pareto-optimal set $\mathcal{A}$, where $i \in [0, \ldots, n_\mathcal{A}$]. Each action is an integer 
$a \in \{0, \ldots, n_{\mathcal{A}}-1\}$, where $n_{\mathcal{A}} \in \mathbb{N}$ represents the number of possibilities to choose an action
from which corresponds to the number of Pareto points available in the safe space $\mathcal{A}$.

\subsection{Designing the reward function}
\label{subsec: designing reward function}
The reward function defines the desired behavior exhibited by the agent. In this context, our objective is to incentivize the agent to minimize lateral and velocity deviations. While designing the reward function, we face a common issue encountered in RL where the agent manipulates the reward system to exploit unintended behaviors, known as the "Cobra Effect" or reward hacking. For instance, an agent chooses to remain at a standstill as it optimizes lateral deviation, contrary to expectations. Works \cite{zarrouki2021weights,zarrouki2020reinforcement} address this issue by introducing a multi-objective Gaussian (MOG) reward function. The latter defines an attraction region and pushes the agent to converge to a certain attraction point where the optimization objectives reach their maximum value:
\begin{equation}
\label{eq:reward_func}
    R(z_{0}, \, \ldots, \, z_{n}) \; = \; A \cdot \exp \left( {- \sum_{k=0}^{n}\frac{(z_{k} - z_{t, k})^{2}}{2\sigma_{k}^{2}}} \right)
\end{equation}
Here, $[z_{0}, \, \ldots, \, z_{n}]$ denote the set of optimization objectives,  $[z_{t,0}, \, \ldots, \, z_{t,n}]$ the respective target values, $[\sigma_{0}, \, \ldots, \, \sigma_{n}]$ the respective standard deviations and $A$ the maximum achievable reward.
In this work, we optimize the lateral- and velocity deviation, i.e., $z_0 = e_\text{lat}$ and $z_1 = e_\text{vel}$ with $z_{t,0} = z_{t,1} = 0$. Note that, $e_\text{lat} = \sqrt{\frac{1}{n} \sum_{i=1}^{n} e_{\text{lat},i}^2}$ and $e_\text{vel} = \sqrt{\frac{1}{n} \sum_{i=1}^{n} e_{\text{vel},i}^2}$ with $n = \frac{T_{sw}}{T_\text{s,sim}}$, i.e. the root mean square (RMS) of the $n$ measured deviations between two parameter switching intervals $T_{sw}$.
To ensure that metrics with varying orders of magnitude contribute proportionately to the overall performance evaluation, we normalize each variable to a certain normalization bounds $[a,b]$. If a metric exceeds these bounds, its value is truncated to the range [0, 1].\\ 
Thanks to the Safe RL design, there is no necessity for special conditions to penalize unintended behaviors that pose safety risks.
\subsection{Defining the observations}
\label{subsec:observation space}
Agent's observations of its environment affect its decision-making process. In our use case, we design the agent to observe the current ego velocity and to assess the performance of its last actions through lateral- and velocity deviations. \\
To enhance the agent's foresight capabilities, we provide the current reference trajectory within the prediction horizon. With this design, we aim for the agent to proactively determine optimal parameters aligned with the specified objectives and according to the anticipated dynamic profiles the MPC should follow.
In our use case, the agent observes future velocity and yaw rate profiles, enabling it to make informed decisions concerning future combined longitudinal and lateral behaviors. This distinction allows the agent to navigate various scenarios, such as high-speed curves and straight paths.
Based on the discussions above, we define a general algorithm that learns and adapts the MPC cost function parameters with Safe RL (Algorithm \ref{alg:RL_SNMPC}). 
\begin{algorithm}[ht]
\caption{Safe RL driven Weights-varying MPC}\label{alg:RL_SNMPC}
\algrenewcommand\algorithmicrequire{\textbf{Init}}
\begin{algorithmic}[1]
\State Set MPC parameters: $T_p$, $T_{s}$, $T_u$, $L1$, $L2$, $W$ and $W_e$
\For{$i \in \left\{1,...,N\right\}$}
    \State Update initial state $\boldsymbol{x}_0$ with current measurements
    \State Update current MPC reference
    \If{\text{switching time (}$i \bmod T_\text{sw} = 0$)}
		\State Generate observations and rewards
		\State Feedforward the current RL policy and get the current optimal RL action $a$
        \State Map the discrete RL action $a$ into the corresponding weights set from the safe space $\mathcal{A}$
		\State Update MPC cost function weights 
        \If{learning-step ($i \bmod n_\text{steps}\cdot T_\text{sw} = 0$)}
        \State Execute RL policy update
        \EndIf
    \EndIf
    \State Solve the MPC problem (Eq.\ref{eq:nominal NMPC problem})
    \State Apply the first control input $\boldsymbol{u}^*_0$ on the real system
\EndFor
\end{algorithmic}
\end{algorithm}
\section{Simulation Results: Safe RL driven Weights-varying NMPC}\label{sec:results}

We evaluate our concept for the combined lateral- and longitudinal motion control of a full-scaled research autonomous vehicle, a Volkswagen T7. We refer to \cite{zarrouki2023stochastic} for further details about the exact problem formulation and the used models. 

\subsection{Training and simulation setup}
The parameters for the MOBO framework have been set as specified in \cref{tab:BO_parameters}. We generated 50 initial samples at random, followed by 400 iterations of BO, leading to 450 iterations in batches of 5. The acquisition function parameters have been set as specified above. We use a separate hypervolume reference point $R_0$ and $R_1$ for each segment group.
\begin{table}[H]
\centering
\caption{Hyperparameters for the MOBO framework}
\label{tab:BO_parameters}
\begin{tabular}{ll}
\toprule
\textbf{Parameter}&\textbf{Value} \\
\midrule
$N_\text{initial}$ & 50 \\
$N_\text{BO}$ & 400 \\
batch size & 5 \\
k & 1.0 \\
$\epsilon$ & 0.9 \\
$R_{0}$ & $[0.5, 0.75]$ \\
$R_{1}$ & $[0.4, 0.9]$ \\
\bottomrule
\end{tabular}
\end{table}

The WMPC configurations are detailed in Table \ref{tab:SNMPC parameter}, while our NMPC implementation specifics can be found in \cite{zarrouki2023stochastic}.\\
The RL agent's training is carried out on an AMD Ryzen 7950X 5.70 GHz CPU using the Proximal Policy Optimization (PPO) algorithm implemented within the Stable Baselines3 framework \cite{stable-baselines3}. Configuration details are outlined in Table \ref{tab:PPO parameter}. The training process in our simulation environment spans approximately 180\unit{\minute} on 16 parallel environments.
\begin{table}[h]
\centering
\caption{Proximal Policy Optimization hyperparameters}
\label{tab:PPO parameter}
\begin{tabular}{ll}
\toprule
\textbf{Parameter}&\textbf{Value} \\
\midrule
Learning rate [initial, final] [$\alpha_\text{i}$, $\alpha_\text{f}$] & [0.005, 0.0001] \\
Learning rate (decay) $\alpha_\text{decay}$&0.4 \\
Policy update every $n_\text{steps}$&512 \\
Discount factor $\gamma$ & 0.8\\
GAE factor $\lambda$ & 0.98\\
Clipping parameter $\epsilon$ & 0.2\\
Entropy coefficient & 0.006 \\
Minibatch size & 4096\\
Training steps $N$ & $1.5\cdot10^{6}$\\
\bottomrule
\end{tabular}
\end{table}
\begin{table}[h]
\centering
\caption{Weights-varying NMPC parameters}
\label{tab:SNMPC parameter}
\begin{tabular}{ll}
\toprule
\textbf{Parameter}&\textbf{Value}\\
\midrule
Simulation sampling time $T_\text{s,sim}$ & 0.02\unit{\second} \\
NMPC discretization time $T_s$ & 0.08\unit{\second} \\
Prediction horizon $T_p$ & 3.04\unit{\second} \\
Weights switching time $T_\text{sw}$ & 1.6\unit{\second} \\
RL look-ahead horizon $T_\text{la}$ & 3.04\unit{\second}\\
Simulation duration per evaluation& 110\unit{\second}  \\
\bottomrule
\end{tabular}
\end{table}
We configure the reward function (Sec.\ref{subsec: designing reward function}) as follows, $A=1$, the normalization bounds for the lateral- and velocity deviations are $[0, 0.4\unit{\meter}]$ and $[0, 1\unit{\meter\per\second}]$ respectively, the reward standard deviations for lateral- and velocity deviations are $\sigma_\text{lat} = 0.1\unit{\meter}$ and $\sigma_\text{vel} = 0.5\unit{\meter\per\second}$, respectively.\\
We train the RL agent on two real-world racetracks: Monteblanco and Modena. Reference trajectories are generated by fully exploiting the vehicle's actuation interface limits in allowed velocity and combined longitudinal and lateral accelerations (Fig.\ref{fig:benchmark_trained_untrained_gg}). Our research vehicle interface allows a maximum velocity of $37.5\unit{\meter\per\second}$ \cite{zarrouki2023stochastic}. After each training episode, the vehicle starts at a random position on one of the training racetracks.
In the following sections, each evaluation consists of $110\unit{\second} = \frac{110}{T_\text{s,sim}} = 5500$ MPC steps.
\subsection{Trajectory following performance}
In Figure \ref{fig:benchmark_trained_untrained_gg}, the closed-loop behavior of both an untrained and a trained RL agent is depicted, evaluated on the Monteblanco racetrack. Notably, both agents perform well with an advantage for the trained agent, especially w.r.t. lateral deviation. During training, the agent learns when to switch to the most optimal weights from the pre-optimized Pareto sets catalog $\mathcal{A}$.
\subsection{Analyzing Reinforcement Learning agent decisions}
Figure \ref{fig:heatmap} illustrates the distribution of policy decisions made by both the untrained and trained agents along the racetrack. Each discrete action corresponds to specific cost function weights in $\mathcal{A}$. The initial DNN weights' randomness influences the decisions of the untrained policy, limiting it to only a subset of the available catalog.
Notably, the trained agent exhibits a successful learning trajectory. It adapts by selecting parameter sets optimized for straight segments (numbers 0 to 10) when driving on long straights. As the track curves, the trained policy gradually transitions to sets optimized for a tradeoff (numbers 11 to 14) and then further to sets specifically tailored for curve segments (numbers 15 to 25).
\subsection{Weights-varying MPC compared to nominal MPC}
\begin{figure}[t!]
    \begin{subfigure}[t]{1\columnwidth}
    \includegraphics[width=1\columnwidth]{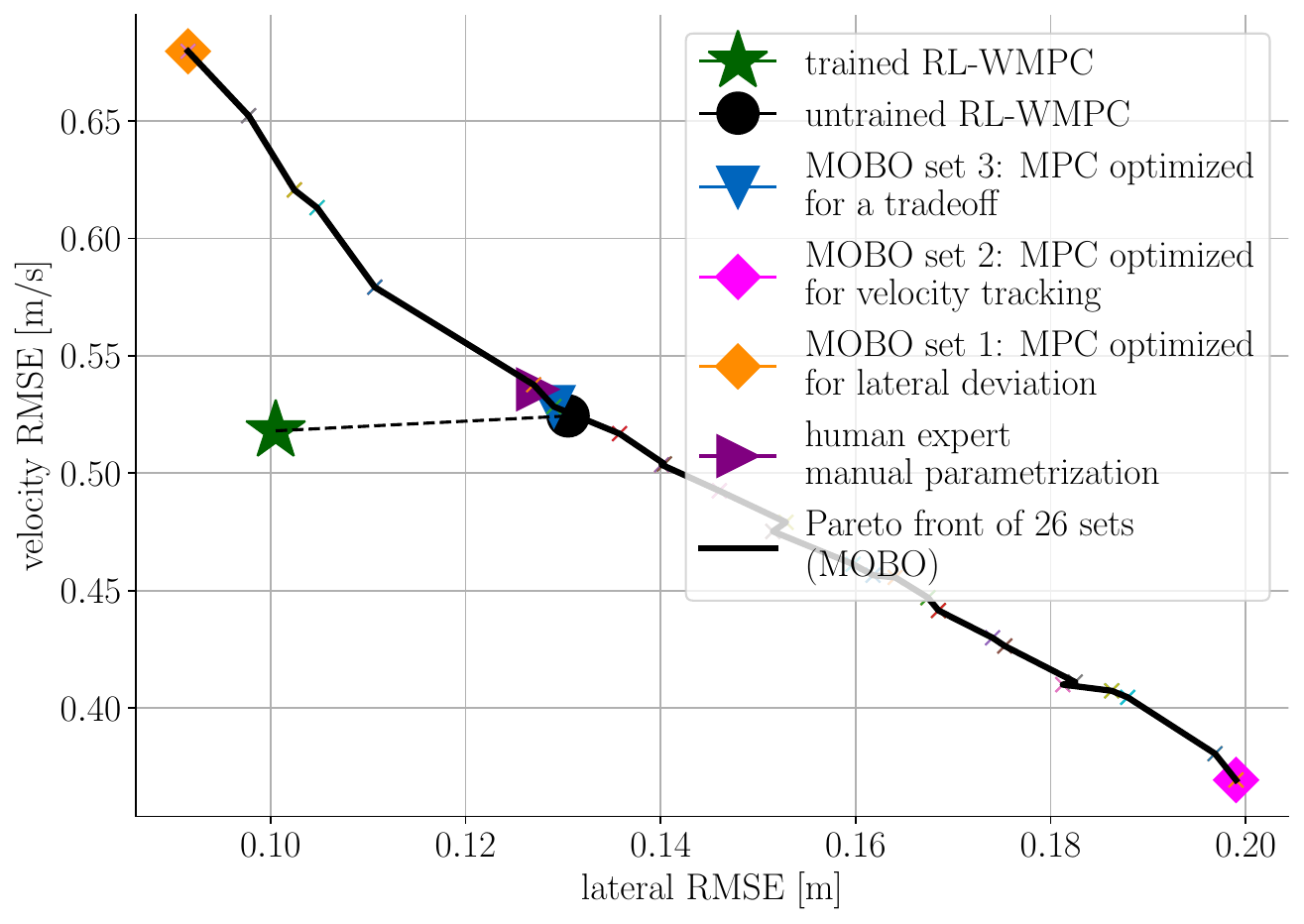}
    \caption{Lateral- to velocity RMSE behavior: a trained RL-WMPC achieves performance beyond the Pareto-front.}
    \label{fig:RMSE_monteblanco}
    \end{subfigure}
    \newline
    \begin{subfigure}[t]{1\columnwidth}
    \includegraphics[width=1\columnwidth]{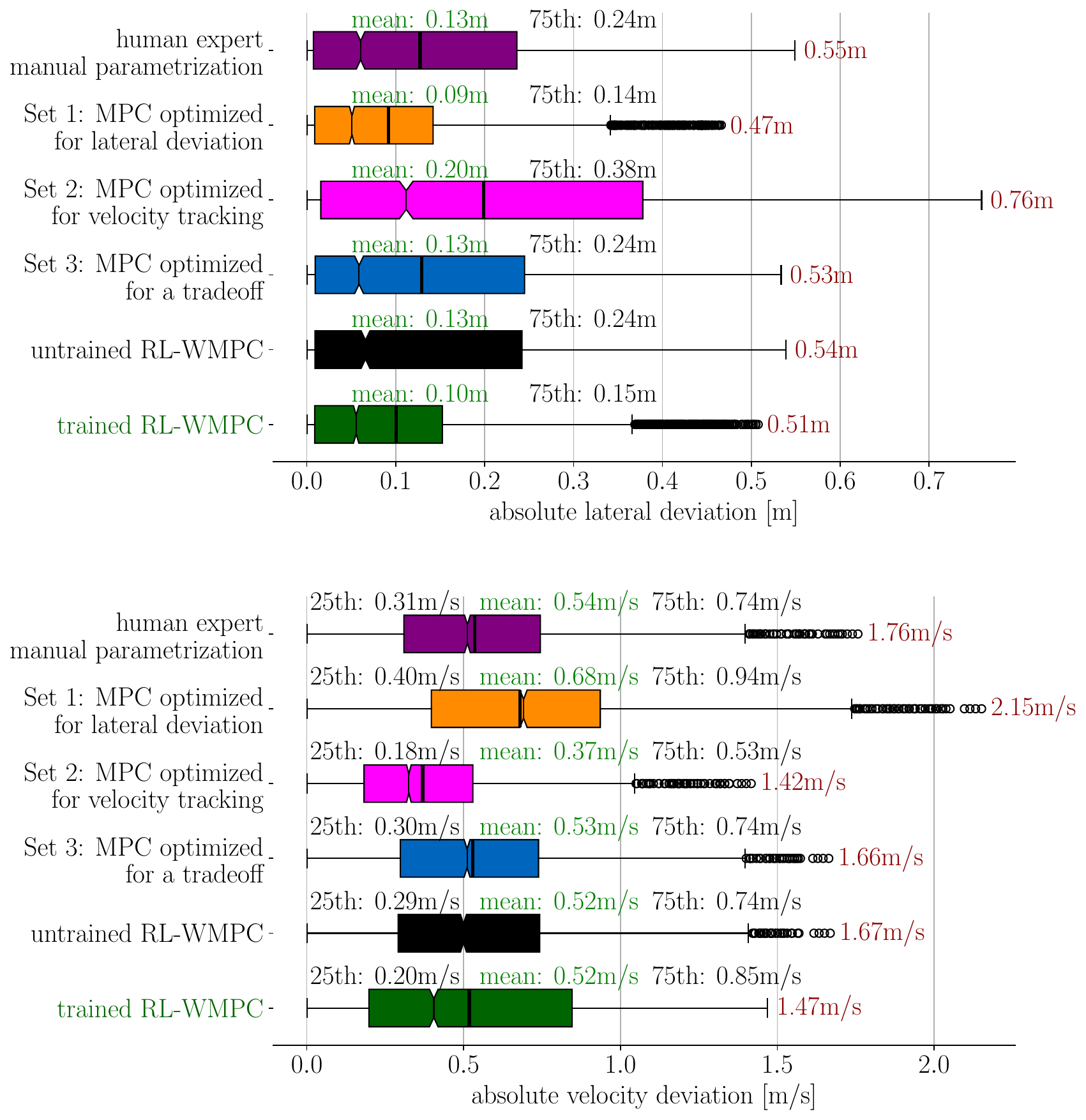}
    \caption{Box plots of the absolute lateral- and velocity deviations.}
    \label{fig:boxplots_monteblanco}
    \end{subfigure}
    
\caption{Benchmark of the closed-loop performance of different settings following an optimal raceline on Monteblanco: RL driven Weights-varying MPC, nominal MPC with manual tuning and with different sets from the Pareto front.}
\label{fig:results_monteblanco}
\end{figure}
In Figure \ref{fig:RMSE_monteblanco}, it is evident that dynamically varying MPC's cost function weights online by switching among different sets of the Pareto front leads to a significantly improved closed-loop performance compared to relying on individual nominal MPCs parameterized with cost function sets from the Pareto front. 
This highlights the efficacy of the RL-WMPC approach in achieving a better tradeoff between predefined control objectives, specifically addressing lateral and velocity deviations.\\
The performance of the untrained RL-WMPC agent closely matches that of a nominal MPC equipped with optimal weights for a tradeoff from the Pareto front and an MPC fine-tuned by a human expert. This proves that our approach ensures a safe RL. \\
In Figure \ref{fig:boxplots_monteblanco}, it is evident that the MPC parameterized by a human expert, the MOBO optimized for a tradeoff, and the untrained RL-WMPC all exhibit similar lateral- and velocity-tracking performances across various metrics. This consistency underscores the safety features inherent in the RL-WMPC approach presented in this work.\\
While a nominal MPC driven by set 1 demonstrates the best lateral behavior, it concurrently exhibits the worst longitudinal behavior. Conversely, the nominal MPC driven by set 2, while excelling in minimizing velocity deviations, experiences the largest lateral deviations.\\
Remarkably, the trained RL-WMPC achieves a performance comparable to set 2 (notably strong in velocity tracking) w.r.t. the maximum velocity deviation. Simultaneously, it exhibits performance comparable to set 1 (particularly effective in lateral tracking) across all lateral deviation metrics.


\subsection{Generalization and robustness}
\label{subsec: generalization}
\begin{figure}[h]
\includegraphics[width=1\columnwidth]{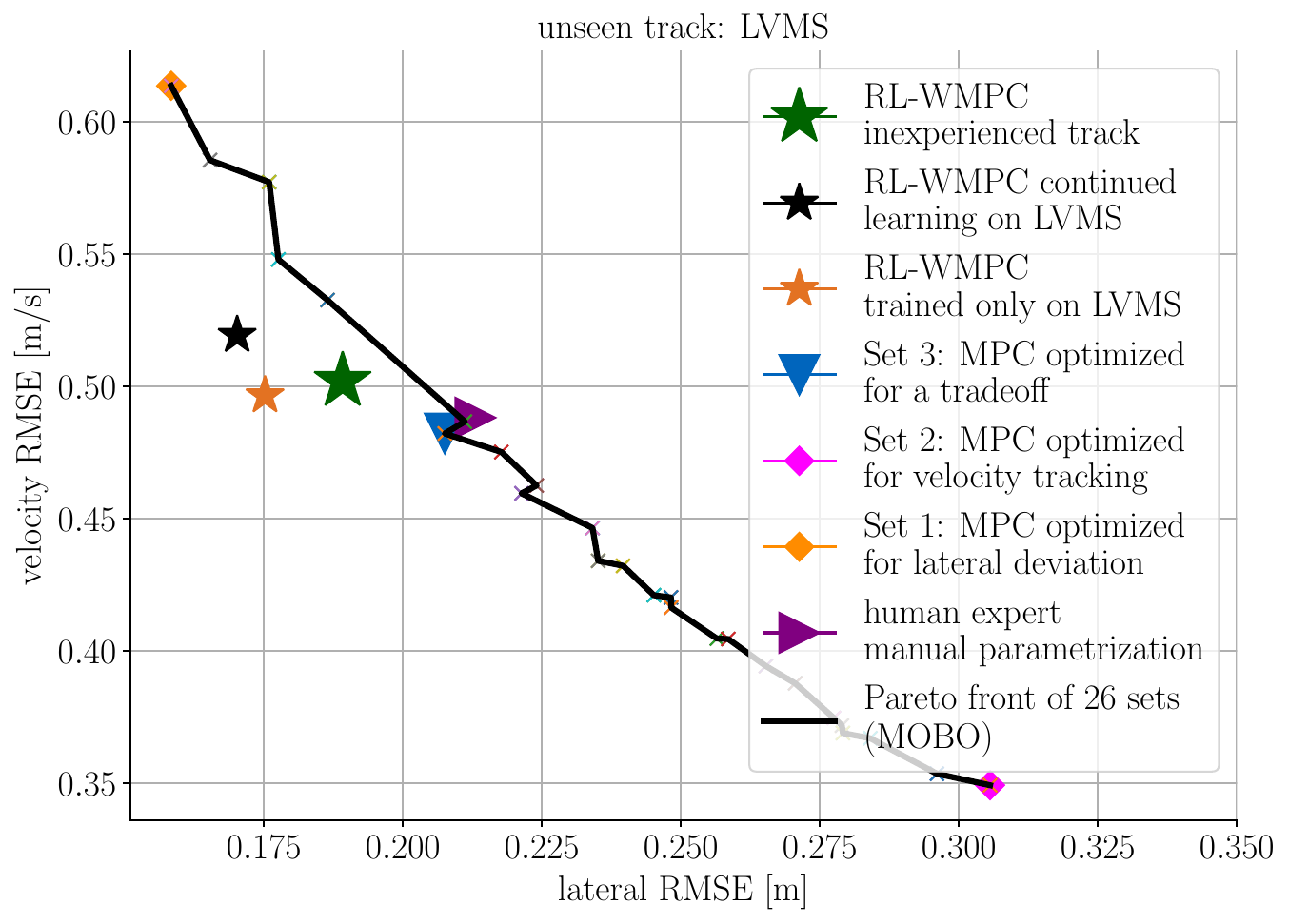}
\caption{Inexperienced Las Vegas Motor Speedway (LVMS): lateral- to velocity RMSE behavior.}
\label{fig:RMSE_LVMS}
\end{figure}

To assess the generalization capabilities of our Safe RL-driven WMPC approach, we conduct evaluations on an unseen track: Las Vegas Motorspeedway (LVMS). This track was not part of the MOBO pre-optimization or RL training phases. In Figure \ref{fig:RMSE_LVMS}, the lateral-to-velocity RMSE behavior is illustrated, similar to the representation in Fig. \ref{fig:RMSE_monteblanco}. Demonstrating consistency, the trained RL-WMPC exhibits performance that consistently surpasses the Pareto-front benchmarks.
The RL-WMPC performance experiences a marginal enhancement in lateral deviation when the agent continues further learning on the new, unexplored track. However, there is a slight degradation in performance concerning velocity RMSE.\\
To comprehensively evaluate RL-WMPC's adaptability to an unfamiliar environment (LVMS), we benchmark its performance against an agent exclusively trained in this specific environment. Notably, the performances exhibit remarkable comparability, with a slight edge observed in favor of the latter.
\section{Conclusions and Future Work}\label{sec:conclusion}
To reduce the manual effort required to identify optimal MPC cost function weights and dynamically adapt them to changing operating conditions, we introduce a two-step automated optimization methodology. Initially, Multiobjective Bayesian Optimization (MOBO) identifies Pareto optimal weight sets. Subsequently, a Look-ahead Deep Reinforcement Learning (RL) agent anticipates upcoming control tasks (reference trajectory) and learns to select the most optimal set of weights from the pre-computed Pareto front. This two-step approach results in a Safe RL driven Weights-varying MPC (WMPC).\\
Experimental findings reveal that the trained RL-WMPC delivers closed-loop performance beyond the Pareto front, i.e., outperforms all the single sets available on the Pareto front in finding a tradeoff between the multi-optimization objectives. 
Experimental results show that even an untrained agent exhibits a Pareto optimal behavior, affirming the validity of our Safe RL training hypothesis.\\ 
However, changes in the MPC prediction model parametrization require a corresponding adjustment in the cost function parametrization. 
In contrast to manual tuning, our approach facilitates automated pre-optimization of the cost function weights coupled with RL training. In future work, we aim to investigate an automated cost function parametrization approach for various prediction model parameters.


\bibliographystyle{IEEEtran}
\bibliography{literatur}

\begin{thebibliography}{10}
\providecommand{\url}[1]{#1}
\csname url@samestyle\endcsname
\providecommand{\newblock}{\relax}
\providecommand{\bibinfo}[2]{#2}
\providecommand{\BIBentrySTDinterwordspacing}{\spaceskip=0pt\relax}
\providecommand{\BIBentryALTinterwordstretchfactor}{4}
\providecommand{\BIBentryALTinterwordspacing}{\spaceskip=\fontdimen2\font plus
\BIBentryALTinterwordstretchfactor\fontdimen3\font minus \fontdimen4\font\relax}
\providecommand{\BIBforeignlanguage}[2]{{%
\expandafter\ifx\csname l@#1\endcsname\relax
\typeout{** WARNING: IEEEtran.bst: No hyphenation pattern has been}%
\typeout{** loaded for the language `#1'. Using the pattern for}%
\typeout{** the default language instead.}%
\else
\language=\csname l@#1\endcsname
\fi
#2}}
\providecommand{\BIBdecl}{\relax}
\BIBdecl

\bibitem{ramasamy2019optimal}
V.~Ramasamy, R.~K. Sidharthan, R.~Kannan, and G.~Muralidharan, ``Optimal tuning of model predictive controller weights using genetic algorithm with interactive decision tree for industrial cement kiln process,'' \emph{Processes}, vol.~7, no.~12, p. 938, 2019.

\bibitem{rodrigues2019tuning}
L.~L. Rodrigues, A.~S. Potts, O.~A. Vilcanqui, and A.~J. Sguarezi~Filho, ``Tuning a model predictive controller for doubly fed induction generator employing a constrained genetic algorithm,'' \emph{IET Electric Power Applications}, vol.~13, no.~6, pp. 812--819, 2019.

\bibitem{kapnopoulos2022cooperative}
A.~Kapnopoulos and A.~Alexandridis, ``A cooperative particle swarm optimization approach for tuning an mpc-based quadrotor trajectory tracking scheme,'' \emph{Aerospace Science and Technology}, vol. 127, p. 107725, 2022.

\bibitem{piga2019performance}
D.~Piga, M.~Forgione, S.~Formentin, and A.~Bemporad, ``Performance-oriented model learning for data-driven mpc design,'' \emph{IEEE control systems letters}, vol.~3, no.~3, pp. 577--582, 2019.

\bibitem{gharib2021multi}
A.~Gharib, D.~Stenger, R.~Ritschel, and R.~Vo{\ss}winkel, ``Multi-objective optimization of a path-following mpc for vehicle guidance: A bayesian optimization approach,'' in \emph{2021 European Control Conference (ECC)}.\hskip 1em plus 0.5em minus 0.4em\relax IEEE, 2021, pp. 2197--2204.

\bibitem{sorourifar2021data}
F.~Sorourifar, G.~Makrygirgos, A.~Mesbah, and J.~A. Paulson, ``A data-driven automatic tuning method for mpc under uncertainty using constrained bayesian optimization,'' \emph{IFAC-PapersOnLine}, vol.~54, no.~3, pp. 243--250, 2021.

\bibitem{paulson2021probabilistically}
J.~A. Paulson, K.~Shao, and A.~Mesbah, ``Probabilistically robust bayesian optimization for data-driven design of arbitrary controllers with gaussian process emulators,'' in \emph{2021 60th IEEE Conference on Decision and Control (CDC)}.\hskip 1em plus 0.5em minus 0.4em\relax IEEE, 2021, pp. 3633--3639.

\bibitem{rupenyan2021performance}
A.~Rupenyan, M.~Khosravi, and J.~Lygeros, ``Performance-based trajectory optimization for path following control using bayesian optimization,'' in \emph{2021 60th IEEE Conference on Decision and Control (CDC)}.\hskip 1em plus 0.5em minus 0.4em\relax IEEE, 2021, pp. 2116--2121.

\bibitem{makrygiorgos2022performance}
G.~Makrygiorgos, A.~D. Bonzanini, V.~Miller, and A.~Mesbah, ``Performance-oriented model learning for control via multi-objective bayesian optimization,'' \emph{Computers \& Chemical Engineering}, vol. 162, p. 107770, 2022.

\bibitem{zarrouki2023adaptive}
B.~Zarrouki, C.~Wang, and J.~Betz, ``Adaptive stochastic nonlinear model predictive control with look-ahead deep reinforcement learning for autonomous vehicle motion control,'' 2023.

\bibitem{bohn2021reinforcement}
E.~B{\o}hn, S.~Moe, S.~Gros, and T.~A. Johansen, ``Reinforcement learning of the prediction horizon in model predictive control,'' \emph{IFAC-PapersOnLine}, vol.~54, no.~6, pp. 314--320, 2021.

\bibitem{bohn2023optimization}
E.~B{\o}hn, S.~Gros, S.~Moe, and T.~A. Johansen, ``Optimization of the model predictive control meta-parameters through reinforcement learning,'' \emph{Engineering Applications of Artificial Intelligence}, vol. 123, p. 106211, 2023.

\bibitem{mehndiratta2018automated}
M.~Mehndiratta, E.~Camci, and E.~Kayacan, ``Automated tuning of nonlinear model predictive controller by reinforcement learning,'' in \emph{2018 IEEE/RSJ International Conference on Intelligent Robots and Systems (IROS)}.\hskip 1em plus 0.5em minus 0.4em\relax IEEE, 2018, pp. 3016--3021.

\bibitem{zarrouki2021weights}
B.~Zarrouki, V.~Kl{\"o}s, N.~Heppner, S.~Schwan, R.~Ritschel, and R.~Vo{\ss}winkel, ``Weights-varying mpc for autonomous vehicle guidance: a deep reinforcement learning approach,'' in \emph{2021 European Control Conference (ECC)}.\hskip 1em plus 0.5em minus 0.4em\relax IEEE, 2021, pp. 119--125.

\bibitem{zarrouki2020reinforcement}
B.~Zarrouki, ``Reinforcement learning of model predictive control parameters for autonomous vehicle guidance,'' \emph{Master's thesis}, 2020.

\bibitem{wabersich2020performance}
K.~P. Wabersich and M.~N. Zeilinger, ``Performance and safety of bayesian model predictive control: Scalable model-based rl with guarantees,'' \emph{arXiv preprint arXiv:2006.03483}, 2020.

\bibitem{berkenkamp2016safe}
F.~Berkenkamp, A.~P. Schoellig, and A.~Krause, ``Safe controller optimization for quadrotors with gaussian processes,'' in \emph{2016 IEEE international conference on robotics and automation (ICRA)}.\hskip 1em plus 0.5em minus 0.4em\relax IEEE, 2016, pp. 491--496.

\bibitem{berkenkamp2017safe}
F.~Berkenkamp, M.~Turchetta, A.~Schoellig, and A.~Krause, ``Safe model-based reinforcement learning with stability guarantees,'' \emph{Advances in neural information processing systems}, vol.~30, 2017.

\bibitem{zarrouki2023stochastic}
B.~Zarrouki, C.~Wang, and J.~Betz, ``A stochastic nonlinear model predictive control with an uncertainty propagation horizon for autonomous vehicle motion control,'' \emph{arXiv preprint arXiv:2310.18753}, 2023.

\bibitem{Emmerich2006}
M.~Emmerich, K.~C. Giannakoglou, and B.~Naujoks, ``{Single- and multiobjective evolutionary optimization assisted by Gaussian random field metamodels},'' \emph{{IEEE Transactions on Evolutionary Computation}}, vol.~10, no.~4, pp. 421--439, 2006.

\bibitem{schulman2017proximal}
J.~Schulman, F.~Wolski, P.~Dhariwal, A.~Radford, and O.~Klimov, ``Proximal policy optimization algorithms,'' 2017.

\bibitem{stable-baselines3}
\BIBentryALTinterwordspacing
A.~Raffin, A.~Hill, A.~Gleave, A.~Kanervisto, M.~Ernestus, and N.~Dormann, ``Stable-baselines3: Reliable reinforcement learning implementations,'' \emph{Journal of Machine Learning Research}, vol.~22, no. 268, pp. 1--8, 2021. [Online]. Available: \url{http://jmlr.org/papers/v22/20-1364.html}
\BIBentrySTDinterwordspacing

\end{thebibliography}

\end{document}